\documentclass[conference]{IEEEtran}
\IEEEoverridecommandlockouts
\usepackage{cite}
\usepackage{amsmath,amssymb,amsfonts}
\usepackage{algorithm}
\usepackage{algorithmicx}
\usepackage{graphicx}
\usepackage{textcomp}
\usepackage{xcolor}
\usepackage{tabularx}
\usepackage{multirow}
\usepackage{booktabs}
\usepackage{algpseudocode}
\usepackage{threeparttable}
\usepackage{url}
\usepackage{newfloat}
\usepackage{listings}
\def\BibTeX{{\rm B\kern-.05em{\sc i\kern-.025em b}\kern-.08em
    T\kern-.1667em\lower.7ex\hbox{E}\kern-.125emX}}
\begin{document}
\makeatletter
\newcommand{\linebreakand}{%
  \end{@IEEEauthorhalign}
  \hfill\mbox{}\par
  \mbox{}\hfill\begin{@IEEEauthorhalign}
}
\makeatother

\title{ChatLogic: Integrating Logic Programming with Large Language Models for Multi-Step Reasoning\\
\thanks{*Corresponding author}
}


\author{\IEEEauthorblockN{Zhongsheng Wang}
\IEEEauthorblockA{\textit{School of Computer Science} \\
\textit{University of Auckland}\\
Auckland, New Zealand \\
zwan516@aucklanduni.ac.nz}
\and
\IEEEauthorblockN{Jiamou Liu*}
\IEEEauthorblockA{\textit{School of Computer Science} \\
\textit{University of Auckland}\\
Auckland, New Zealand \\
jiamou.liu@auckland.ac.nz}
\and
\IEEEauthorblockN{Qiming Bao}
\IEEEauthorblockA{\textit{School of Computer Science} \\
\textit{University of Auckland}\\
Auckland, New Zealand \\
qbao775@aucklanduni.ac.nz}
\linebreakand 
\IEEEauthorblockN{Hongfei Rong}
\IEEEauthorblockA{\textit{School of Computer Science} \\
\textit{University of Auckland}\\
Auckland, New Zealand \\
hron635@aucklanduni.ac.nz}
\and
\IEEEauthorblockN{Jingfeng Zhang}
\IEEEauthorblockA{\textit{School of Computer Science} \\
\textit{University of Auckland}\\
Auckland, New Zealand \\
jingfeng.zhang@auckland.ac.nz}
}

\maketitle

\begin{abstract}
Large language models (LLMs) such as ChatGPT and GPT-4 have demonstrated impressive capabilities in various generative tasks. However, their performance is often hampered by limitations in accessing and leveraging long-term memory, leading to specific vulnerabilities and biases, especially during long interactions. This paper introduces ChatLogic, an innovative framework specifically targeted at LLM reasoning tasks that can enhance the performance of LLMs in multi-step deductive reasoning tasks by integrating logic programming. In ChatLogic, the language model plays a central role, acting as a controller and participating in every system operation stage. We propose a novel method of converting logic problems into symbolic integration with an inference engine. This approach leverages large language models' situational understanding and imitation skills and uses symbolic memory to enhance multi-step deductive reasoning capabilities. Our results show that the ChatLogic framework significantly improves the multi-step reasoning capabilities of LLMs. The source code and data are available at \url{https://github.com/Strong-AI-Lab/ChatLogic}.
\end{abstract}

\begin{IEEEkeywords}
deductive reasoning, multi-step reasoning, large language models
\end{IEEEkeywords}

\section{Introduction}

Recent advancements in large language models (LLMs) such as ChatGPT-3.5, GPT-4 \cite{OpenAI2023GPT4TR}, and Llama2 \cite{touvron2023llama} have significantly enhanced their capabilities in various industries, proving invaluable in solving complex real-world problems. These models revolutionize sectors like customer service, healthcare, and education through their nuanced contextual comprehension and advanced conversational abilities. However, LLMs face notable challenges in multi-step logic reasoning tasks.


While these models excel in content generation, they consistently struggle to produce coherent responses in tasks requiring multi-step reasoning. Their training methodology, primarily based on the `next-token prediction' approach, limits their ability to apply logical rules and deep contextual understanding essential for such tasks. For example, Figure \ref{fig:my_label} shows how to let LLMs find a reasonable explanation path as the judgment result of the problem in the known randomly disrupted proposition sequence. This represents a critical area for improvement in current LLMs.

Further complicating this issue is the inherent token limitation of LLMs, which becomes apparent in continual dialogues \cite{thirunavukarasu2023large}. The token caps in models like GPT-3.5 and GPT-4, while extendable through engineering prompts or technologies like Recursive Model Training \cite{bulatov2023scaling}, still pose a significant constraint. This limitation is particularly pronounced in multi-turn conversations, a common feature in multi-step logic reasoning tasks.

\begin{figure}[h] 
    \centering
    \includegraphics[width=0.8\columnwidth]{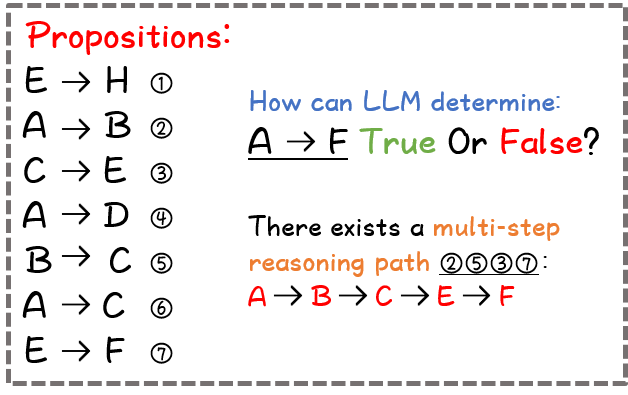} 
    \caption{Demo illustrating how LLMs can effectively identify and follow correct and logical reasoning paths to solve complex multi-step reasoning problems. In this instance, our objective is to let LLMs recognize the presence of an established path, ABCEF, thereby enabling them to accurately deduce that the statement `A infers F' is true.}
    \label{fig:my_label}
\end{figure}

To address these limitations, innovative approaches such as external memory augmentation are being explored \cite{zhong2022training}. This method involves integrating LLMs with extensive databases to enhance their reasoning capabilities \cite{borgeaud2022improving}. However, this integration brings its own challenges, such as the potential embedding of biases from the retrieval models into the LLMs \cite{khattab2022demonstrate}, which could affect their accuracy and stability.

Our work introduces ChatLogic, a framework that augments LLMs with a logical reasoning engine to enhance their inferential capabilities. We have innovatively implemented a `Mix-shot Chain of Thought' technique in this framework. This approach significantly enhances the performance of LLMs by combining various prompt engineering methods. Mix-shot CoT efficiently guides the model through logical reasoning steps, achieving improved problem-solving with minimal resource consumption. ChatLogic is designed to be compatible with existing LLMs, significantly increasing their accuracy, especially in high-precision scenarios. The framework orchestrates the functioning of an LLM, enabling it to efficiently generate responses across various tasks.

At the heart of ChatLogic is the transformation of natural language into logical symbols, a process executed through pyDatalog. The primary objective of ChatLogic is to reinforce the stability of the reasoning process, ensuring that LLMs can handle intricate reasoning tasks with enhanced reliability and precision.  The main characteristics of our framework are summarized below:

\begin{itemize}

\item ChatLogic, by combining LLMs with pyDatalog, translates natural language queries into logic programs, enhancing inference accuracy. This improvement is notably evident in multi-step reasoning tasks, as demonstrated on datasets such as PARARULE-Plus\footnote{\url{https://huggingface.co/datasets/qbao775/PARARULE-Plus}}, CONCEPTRULES V1, and CONCEPTRULES V2.


\item ChatLogic mitigates information loss, effectively addressing the long sequence limitation prevalent in adopting LLMs for multi-step reasoning tasks.


\item ChatLogic incorporates automated enhancements for logic program execution, including a syntax correction module. This module refines a logic program by learning from previous executions, significantly improving the practical application and effectiveness of the generated code.


\end{itemize}

\section{Related Work}

\subsection{LLMs Reasoning:}
LLMs are considered to have reasoning abilities similar to human cognition \cite{huang2022towards}. Despite facing challenges in multi-step logical tasks involving contemporary information or complex logical sequences \cite{creswell2022selection}, emerging approaches like self-consistency \cite{wang2022self} show promise in enhancing performance, particularly in areas such as arithmetic and common sense reasoning. The effectiveness of causal reasoning pathways \cite{creswell2022faithful} is also crucial, ensuring that the output of LLMs is accurate but also transparent and verifiable. However, the most impactful method is the Chain of Thought (CoT) \cite{wei2022chain}, which reveals the intermediate reasoning steps used by these models in problem-solving, allowing for continual self-correction of logical thinking, thereby greatly enhancing the rationality of their reasoning capabilities. But they all more or less exposed the weakness of extracting effective content from long and disordered information. We are trying to find ways to improve this shortcoming effectively.

\subsection{LLMs Code Generation:}
LLMs have demonstrated the ability to generate code in various programming languages to meet users' specific needs \cite{yang2023coupling}. However, how to directly apply the generated code to actual environments remains an issue to be resolved. In terms of optimization, the SELF-DEBUGGING approach \cite{chen2023teaching} leads the post-code generation phase. It endows LLMs with the ability to debug their output, reinforcing the theme of continual refinement in the generated code. LOGIC-LM \cite{pan2023logic} creates a deterministic symbolic solver that expresses reasoning in a specific symbolic format to obtain actual results. The existing flaw we hope to improve is that while LLMs may not fully understand demo samples or generate "fantasies" that deviate from reality due to lack of pyDatalog knowledge in the pre-training data, they can still simply "imitate" Produce high-precision output. Our ultimate goal is to generate code that perfectly meets the requirements, can be deployed directly locally, and get inference results through the user's computer with a basic Python language environment installed.

\subsection{LLMs Prompt Engineering:}
Prompt engineering in LLMs functions akin to psychological suggestions, guiding the model towards specific predictions \cite{wang2020generalizing}.
Few-shot learning emphasizes training models with the least labelled data for optimized task performance. Notably, models like GPT-3 can handle tasks with a few examples, comparable to fine-tuned models \cite{brown2020language}. Enhanced by prompt engineering, their reasoning capabilities are magnified. Zero-sample prompt\cite{reynolds2021prompt} completely relies on the model's huge intrinsic knowledge and training corpus and assumes full responsibility for solving the problem. This means that we do not need to make any downstream task-related fine-tuning of LLMs; we only provide task content and expect good performance of LLMs. Surprisingly, although guidance is limited, in many cases, it often produces results that exceed expectations. In addition, zero-sample CoT \cite{kojima2022large} is also considered to be the best inference prompt currently; by using a simple prompt: `Let's think step by step', the specific prompt and the corresponding two-stage key answer extraction prompt technology has significantly improved multiple inference-related zero-shot tasks. surpassed previous zero-shot learning.

Although not every situation using Zero-shot CoT results in optimal content output, in most cases targeting specific downstream tasks, LLMs have shown potential in the current field to effectively perform tasks using small sample techniques and combining external enhancement symbols \cite{song2022llm}. In ChatLogic, we create independent prompt templates for different links in the framework and call them independently. Preliminary results suggest a promising direction by emphasizing LLMs' inherent reasoning capabilities and combining them with basic symbolic rules.

\section{Task Definition}

In our experience with advanced LLMs such as Llama2 and GPT-4, we've noticed their impressive ability to convert text into formal structures like math equations \cite{he2023solving} and programming languages \cite{10.1145/3491101.3519665}. However, these models sometimes struggle with complex, multi-step reasoning tasks. The challenge escalates as the reasoning depth increases, and LLMs often miss key reasoning steps.

Acknowledging these characteristics, our primary goal is to boost the capability of LLMs to effectively represent problems in logic programming languages, particularly pyDatalog. This Python library integrates the logic programming paradigm and is particularly useful for declarative reasoning and complex querying. It allows for sophisticated rule-based logic and inference to be seamlessly incorporated into Python applications, enhancing their capabilities for decision-making processes. 

We specify the multi-step deductive reasoning problem as Fact, Rule, and Query. The data sets used in the paper experiments all have this format:

\begin{itemize}

\item Fact: A Fact $F=\{f_1, f_2 \cdots, f_n\}$ is a sequence of sentences, each declarative sentence $f_i$ having a subject, a predicate verb, and an object, like `Bob is poor' and `Dogs like cats'. Predicates have negative expressions.

\item Rule: A Rule $R=\{r_1, r_2 \cdots, r_n\}$ is a sequence of declarative sentences with conditional judgment, including initial features and inferential features, like `If someone is poor then they are bad.'

\item Query: A Query $Q$ is also a declarative sentence consistent with the sentence expression format.

\end{itemize}

To accomplish this goal, we aim to address the following two subtasks.

\subsection{Augmenting the Inferential Abilities of LLMs}

Our preliminary objective is to exploit LLMs' one-shot generalization and zero-shot thinking capabilities $C$ (the collective name for both technologies). To achieve this, we aim to familiarize the model with the intricacies of symbolic language, specifically utilizing pyDatalog. We introduce them to this language through meticulously crafted examples that cover all potential edge cases. The structured pyDatalog syntax and these detailed prompts example set $S$ are provided to improve skill C and are essential to guide LLMs in understanding and handling multi-step inference.

$$pyDatalogCode = C(LLM(F, R, Q), S)$$

The code generated with high precision is executed by the local compiler to obtain accurate multi-step inference results.

$$Result = LocalExecution(pyDatalogCode)$$

\subsection{Amplifying the Executability of Automated Code Generation Processes} 

The translation of LLMs from text to code \cite{5387226} is often imperfect on the first try and may contain errors. We aim to design a specific module $M$ in ChatLogic that ensures high-accuracy alignment of natural language and translation codes. The generated code should be easily executable locally, producing the desired results directly. Essentially, it further improves code generation quality so that it can be performed accurately and reasonably.

$$pyDatalogCode = M(C(LLM(F, R, Q), S))$$

\begin{figure*}[h] 
  \centering
  \includegraphics[width=\textwidth]{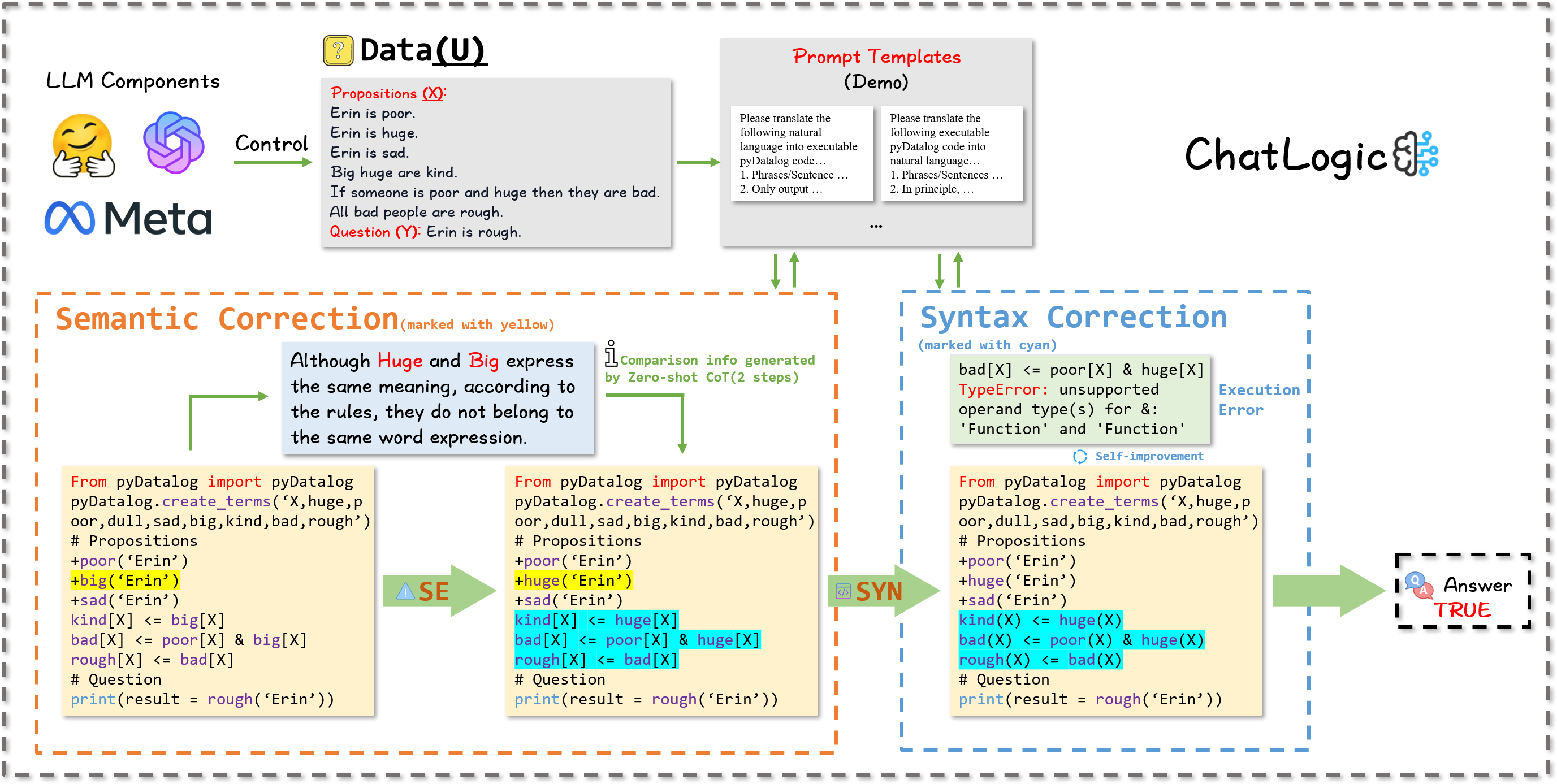}
  \caption{ChatLogic containing more details uses LLMs as controllers, calls appropriate demonstration examples from Prompt Templates, guides the two modules of semantic correction(SE) and syntax correction(SYN) to output correct code, and produces execution results. This excerpts a specific question in PARARULE-Plus and the code generation process. The yellow portion represents the achievements of SE, and the cyan portion represents SYN.}
  \label{fig:Detail}
\end{figure*}

\section{ChatLogic}

This section provides a detailed overview of the ChatLogic framework, particularly emphasising the finer details of its constituent parts and our innovative strategy regarding Mix-shot CoT.

\subsection{Framework Overview}

The ChatLogic framework comprises four primary phases: input processing, Semantic Correction, Syntax Correction, and local execution response, as meticulously illustrated in Figure \ref{fig:Detail}. The entire process, from problem input to result output, is depicted in the image as a demonstration. It is worth noting that the initial version of the logic code generated by LLMs at the beginning, after continuous revision through two or more iterations within two modules (semantic and syntactic correction), as a direct code solution for multi-step reasoning Executability is significantly improved, resulting in more precise results. Moreover, the enhanced refinement of this code substantially bolsters its executability, contributing to improved system performance and accuracy. The blue and green boxes are the intermediate self-correction processes that appear in the semantic correction and syntax correction modules and are generated correspondingly by the ChatLogic framework during the experiment.

Algorithm \ref{alg:algorithm} delves deeper into the comprehensive algorithmic process for querying response data in ChatLogic. Apart from the locally executed part, all sub-tasks within ChatLogic are controlled and driven by LLMs acting as components. It consists of two loops, each corresponding to the content of two correction phases. We observe that LLMs excel at semantic corrections, and with limited modifications, correct text translations can be achieved. In lines 5 and 6 of the code, we employed zero-shot CoT to assist in determining the textual similarity of two propositions. Based on the judgment, we update the `DifferentFlag' label, which influences the progression of the loop process of `Semantic Correction'. However, syntax corrections are unreliable; they may get stuck in an infinite loop, repeatedly performing meaningless tasks. To address this issue, we consider introducing an upper loop limit. Although this somewhat diminishes ChatLogic's inferential capabilities, it significantly enhances the framework's robustness, making it better suited for multi-step deductive reasoning tasks.

\begin{algorithm*}[!h]
\caption{The Algorithm of ChatLogic}
\label{alg:algorithm}
\textbf{Input}: 
U $\leftarrow$ Rules supplemented based on the close-world assumption  \\
X $\leftarrow$ Proposition group (contains facts and rules) \\
Y $\leftarrow$ Question \\
\textbf{Output}: TRUE/FALSE (Answer to Y given U, X)

\begin{algorithmic}[1] 

    \State DifferentFlag = TRUE
    \Statex \Comment{Semantic Correction}
    \While{DifferentFlag}
        \State Code $\leftarrow$ PropositionTransformation($X, Y, U$) 
        \Comment{Generate logic program based on close-world assumption}
        \State RevProposition $\leftarrow$ ReverseTransformation(Code, U) \Comment{Convert code back to natural language}
        
        \State DifferentInfo $\leftarrow$ TextComparison(($X, Y$), RevProposition)
        \State DifferentFlag $\leftarrow$ JudgeInfo(DifferentInfo)
        \Comment{Determining semantic similarity status with zero-shot CoT in 2 steps}
    \EndWhile
    \Statex \Comment{Syntax Correction}
    \State ExecutionError = NULL \Comment{Execution result record}
    \While{Code cannot be executed}
        \State Code $\leftarrow$ CodeImprovement(Code, $\text{ExecutionError}$) \Comment{Improve code based on error info}
        \If{Running Time Overflow}
            \State Terminate WHILE Loop
        \EndIf
    \EndWhile

\State
\Return CodeExecution(Code) \Comment{Get results by executing pyDatalog code locally}
\end{algorithmic}

\end{algorithm*}

\subsection{Mix-shot CoT}

Our innovative mix-shot CoT (Chain of Thought) approach represents a groundbreaking hybrid methodology, blending the strengths of zero-shot CoT and one-shot learning to create a more versatile and effective learning paradigm for language models. As mentioned before, zero-shot CoT uses LLMs' generative capabilities and chain thinking patterns to complete natural language tasks without targeted training or fine-tuning. One-shot learning provides a well-written task completion sample manually to guide LLMs to imitate the same workflow to achieve high-precision task completion. At its core, mix-shot CoT leverages the language model's innate ability for autonomous sub-task identification; a character follows established patterns accurately es it with the precision of one-shot learning through strategically chosen demonstration examples. This dual approach allows for dynamic adaptation to various tasks' complexity and specific requirements. Mix-shot CoT guides the model using high-quality demonstration examples as templates, enhancing its precision and contextual depth for tasks demanding high accuracy and nuanced understanding, like in the method mentioned in this paper to convert natural language questions into logical codes. In addition, in scenarios that require more extensive analysis, such as text similarity comparison between generated propositions and original propositions in ChatLogic, LLMs are given greater autonomy to leverage their analytical capabilities and generate innovative solutions, thereby developing their ability to navigate massive amounts of information and generate unique insights. In addition, using the two-step calling process of zero-shot cot, during the second call to LLMs, we will extract valid key information from the first step as a status label to guide the normal operation of the entire framework.

Moreover, mix-shot CoT is designed to cultivate an adaptable learning process in language models, harmoniously combining structured guidance with the freedom of exploration. This flexibility is crucial in enabling the model to follow established patterns, innovate, and adapt to diverse tasks. 

In our comparative analysis, as depicted in Table \ref{tab:my-table}, our mix-shot CoT methodology showcases a considerable leap in performance by striking an optimal balance between the grounded precision of one-shot learning and the generative flexibility of zero-shot CoT. Our approach reduces hallucination significantly and boasts the highest task-specific accuracy due to its judicious use of high-quality demonstration examples. While it does not entirely eliminate the need for demonstrations like zero-shot CoT, nor does it match the minimal hallucination levels of one-shot learning, the mix-shot CoT's enhanced adaptability and efficiency make it a powerful tool in the realm of prompt engineering. By acknowledging the limitations of requiring some demonstrations and not being as inherently scalable as zero-shot CoT, our mix-shot CoT nonetheless stands out for its pragmatic effectiveness in real-world applications where precision and adaptability are paramount.

\begin{table*}[htbp]
\centering
\caption{Comparison among different methods for prompt engineering}
\label{tab:my-table}
\renewcommand{\arraystretch}{1.3}
\begin{tabular}{c|p{4.7cm}|p{4.7cm}|p{4.7cm}}
\hline
\multirow{2}{*}{Attributes} & \multicolumn{3}{c}{Prompt Engineering Methods} \\
\cline{2-4}
                            & One-shot Learning \cite{brown2020language} & Zero-shot CoT \cite{kojima2022large} & Mix-shot CoT (Ours) \\
\hline
Hallucination                  & Least hallucination & High hallucination & Less hallucination \\
Accuracy                    & Poor performance & Relatively high accuracy & Highest accurate response to task \\
Expansibility               & Different demos needed for different tasks & No demo needed, only simple guidance required & Fewer demos, and improved performance than fewer shots \\
\hline

\end{tabular}
\end{table*}

\section{Evaluation}

In this section, we conduct experiments to evaluate the effectiveness of ChatLogic-augmented LLMs. Our experimental results show that \textbf{ChatLogic+LLMs} significantly outperform baseline LLMs, highlighting the advantages of using logical symbols to enhance the multi-step reasoning capabilities of LLMs.    

\subsection{Datasets and Metrics}

All reasoning questions in PARARULE-Plus adhere to the closed-world assumption, totalling approximately 400,000 samples. It features linguistic information in two contextual scenarios: People and Animal. For this dataset, we conducted our experiment by randomly selecting 50 instances from each depth level in both the Animal and People categories, combining them to form a set of 100 test cases for each depth level, ranging from \textit{Depth=2} to \textit{Depth=5}.

In addition to PARARULE-Plus, we also incorporate the CONCEPTRULES V1\footnote{\url{https://bit.ly/3uVemXG}} and CONCEPTRULES V2\footnote{\url{https://bit.ly/3PApIIB}} datasets in our study. These datasets contain samples that require multi-step reasoning, with depths up to 3, making them suitable for evaluating models' capabilities in complex reasoning tasks. They are available in both simplified and full versions. For each version of the CONCEPTRULES datasets, we initially consolidated all data from the train, test, and dev sets into a single pool. We randomly sampled 100 instances from this unified dataset for our tests.

In our experiments with ChatGPT, GPT-4, and Llama 2-7B, we aimed to establish a baseline for the reasoning capabilities of LLMs as documented in the literature\cite{hu2023chatdb}. A significant part of our study was the implementation of ChatLogic, a framework designed to enhance the accuracy of these models' reasoning. This involved testing configurations like ChatGPT vs. ChatLogic (ChatGPT) across uniform scenarios using instances from the PARARULE-Plus and CONCEPTRULES datasets. The crux of our hypothesis is that if models augmented with ChatLogic demonstrate improved reasoning performance across various difficulty levels compared to their baseline, it would be a strong indicator of ChatLogic's effectiveness. Such results would suggest that ChatLogic could be a valuable addition to the field of artificial intelligence and natural language processing, affirming its utility in advancing the reasoning capabilities of LLMs.

\subsection{LLMs Configuration}

In the ChatLogic framework invocation, ensuring the controllability of the text is paramount. For ChatGPT and GPT-4, the model invocation versions are respectively ``\textbf{gpt-3.5-turbo}'' and ``\textbf{gpt-4}'', and both have the hyperparameter temperature set to 0 so that we can have more control over what the LLM outputs. For Llama 2-7B, using an NVIDIA GeForce RTX 3090 with 24GB of memory, we utilized the transformer-based version \textbf{meta-llama/Llama-2-7b-chat-hf} \footnote{\url{https://huggingface.co/meta-llama/Llama-2-7b-chat-hf}} provided by Huggingface. 

\subsection{Intermediate Process}

\begin{figure*}[h] 
  \centering
  \includegraphics[width=0.8\textwidth]{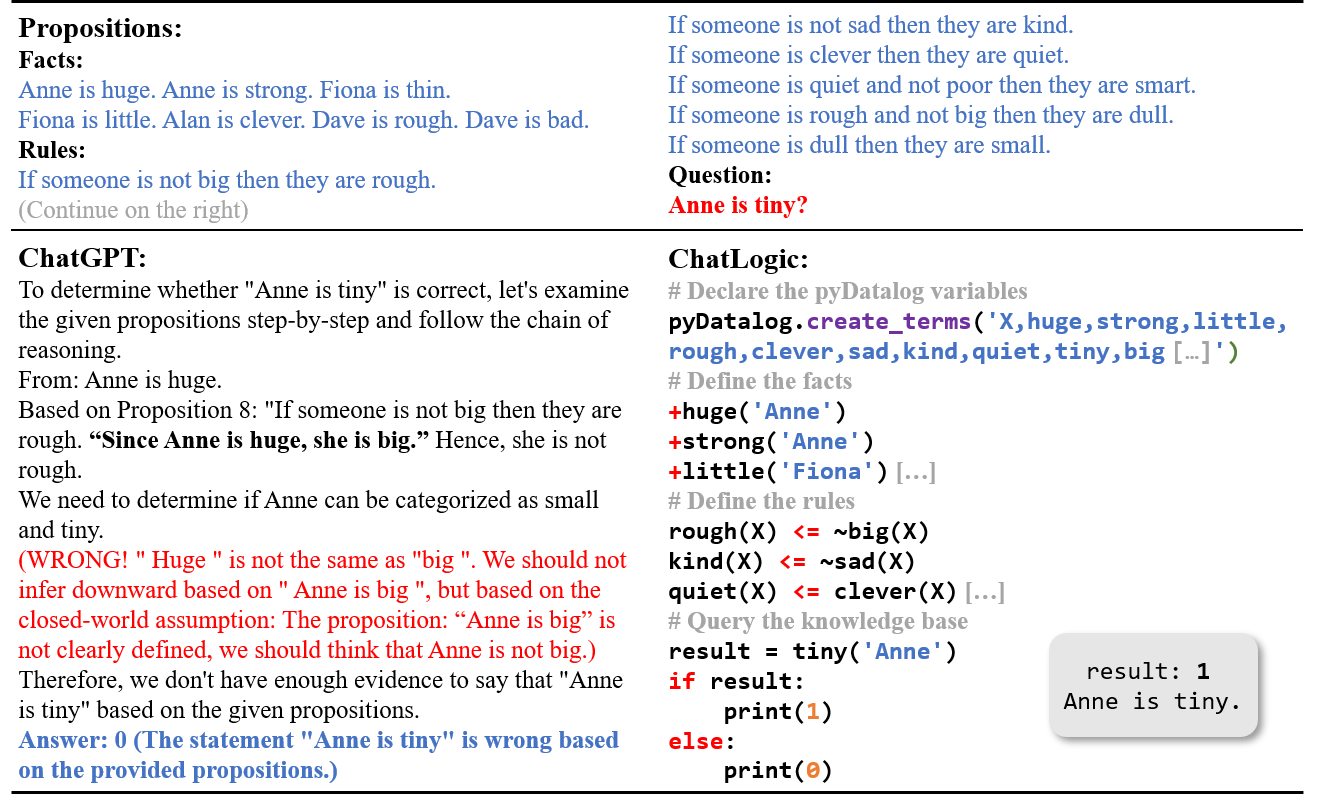}
  \caption{Comparison based on the PARARULE-Plus dataset shows that while ChatGPT, even with CoT reasoning, often leads to incorrect inferences, the ChatLogic framework (also driven by ChatGPT) in most cases accurately generates pyDatalog code, highlighting its more reliable reasoning proficiency.}
  \label{fig:Comparison}

\end{figure*}

Figure \ref{fig:Comparison} provides an illustrative comparison between two approaches for handling the PARARULE-Plus dataset. On the left, we can observe the inference process generated directly by ChatGPT. While it appears initially logical, upon closer examination, it becomes apparent that this inference process contains numerous logical inconsistencies.

In contrast, on the right side of the figure, we showcase the enhanced performance of ChatGPT when coupled with the ChatLogic framework. This presentation includes only a selected excerpt of the final generated code. When executed locally, this integration produces remarkably precise inference results, addressing the logical inconsistencies encountered in the unassisted ChatGPT inference process.

It is worth noting that after experiments, we found that before calling the text data in the PARARULE-Plus data set, LLMs can not complete the missing information for the relevant text content based on the closed-world assumption, resulting in a significant decrease in reasoning efficiency. For example, the beginning of a multi-step deductive reasoning chain process must be the existing attributes of an object, which we call "meta attributes". Meta attributes must be attributes the object possesses rather than attributes derived through inference. This is missing in the data set based on the closed-world assumption mentioned in the paper, so it needs to be supplemented manually. To avoid this situation, we built a script to extract the subject and attribute affiliations in the text through Named Entity Recognition (NER) and complete the missing information in the proposition in the artificially synthesized natural language format. Then, use the ChatLogic framework to enhance text reasoning on LLMs. For example, in the \textbf{`Rules'} in Figure \ref{fig:Comparison}, through observation, we find that no sentence can deduce that someone has the attribute of \textbf{`big'}, so this attribute must appear in the definition of character attributes. If we do not see tasks with \textbf{`big'} attributes in \textbf{`Facts'}, then we default that they do not have \textbf{`big'} attributes, which requires us to manually add them to \textbf{`Facts'}. This part also belongs to the specific content represented by \textbf{U} in Algorithm \ref{alg:algorithm} and Figure \ref{fig:Detail}. The same approach was not utilized for the two distinct versions (simplified \& full) of each of the CONCEPTRULES datasets (V1 \& V2), as these versions did not require the supplementation of additional information.

\begin{table*}[!hbtp]
\centering  
\caption{Accuracy Comparison on the PARARULE-Plus Dataset}
\begin{tabular}{@{}llllllll@{}}
\toprule
Model                    & Method & Depth=2 & Depth=3 & Depth=4 & Depth=5 & Total \\ 
\midrule 
\multirow{3}{*}{GPT-3.5} & Base  & 0.4 & 0.34 & 0.32 & 0.3 & 0.344 \\
                        & Zero-shot CoT  & 0.42 & 0.42 & 0.41 & 0.3 & 0.3875 \\
                        & ChatLogic  & \textbf{0.49} & \textbf{0.56} & \textbf{0.65} & \textbf{0.41} & \textbf{0.5275} \\

\midrule
\multirow{3}{*}{GPT-4}  &  Base & 0.65 & 0.75 & 0.42 & 0.4 & 0.555 \\
                        & Zero-shot CoT & \textbf{0.72} & 0.72 & 0.62 & \textbf{0.7} & 0.69 \\
                        & ChatLogic & \textbf{0.72} & \textbf{0.8} & \textbf{0.7} & \textbf{0.7} & \textbf{0.73} \\
                        
\midrule
\multirow{3}{*}{Llama 2-7B} & Base & 0.11 & 0.06 & 0.01 & 0.01 & 0.0475 \\  
                           & Zero-shot CoT & 0.15 & \textbf{0.13} & 0.08 & 0.06 & 0.105 \\
                           & ChatLogic & \textbf{0.2} & \textbf{0.13} & \textbf{0.22} & \textbf{0.18} & \textbf{0.1825} \\
\bottomrule
\label{tab:1}
\end{tabular}

\textit{Note: This table evaluates accuracy on the PARARULE-Plus dataset, with a score of 1 indicating perfect accuracy. It includes comparisons with the `Base' and `Zero-shot CoT' methods. The `ChatLogic' framework generally surpasses other models, highlighting its superior effectiveness with LLMs. For each depth level, the best performances are highlighted in bold.}
\end{table*}

\begin{table}[!htbp] 
\centering
\caption{Accuracy Comparison on CONCEPTRULES V1 and V2 Datasets}
\begin{tabularx}{\columnwidth}{llllll} 
\toprule
\multirow{2}{*}{\scriptsize Model} & \multirow{2}{*}{\scriptsize Method} & \multicolumn{2}{l}{\scriptsize CONCEPTRULES V1} & \multicolumn{2}{l}{\scriptsize CONCEPTRULES V2} \\
                            &               & \scriptsize simplified & \scriptsize full & \scriptsize simplified & \scriptsize full \\ \midrule
\multirow{3}{*}{\scriptsize GPT-3.5} & \scriptsize Base          & 0.57      &  0.55    & 0.5        &  0.51    \\
                         & \scriptsize Zero-shot CoT & 0.63       &  0.51    & 0.7       &  0.67    \\
                         & \scriptsize ChatLogic     & \textbf{0.69}        &  \textbf{0.67}    & \textbf{0.79}       &  \textbf{0.74}     \\ \midrule
\multirow{3}{*}{\scriptsize GPT-4}   & \scriptsize Base          &  0.95          &  0.94    &    0.89        &  0.86    \\
                         & \scriptsize Zero-shot CoT &  \textbf{0.96}          &  \textbf{0.97}    &    \textbf{0.95}        &  \textbf{0.94}    \\
                         & \scriptsize ChatLogic     &  \textbf{0.96}           &  0.96     &   0.94          &  \textbf{0.94}     \\ \midrule
\multirow{3}{*}{\scriptsize Llama 2-7B} & \scriptsize Base       & 0.32          & 0.29     & 0.31          &  0.24    \\
                            & \scriptsize Zero-shot CoT &  0.42        &  0.41    &   0.33         &  0.3    \\
                            & \scriptsize ChatLogic   &  \textbf{0.48}          &  \textbf{0.49}    &   \textbf{0.37}         &   \textbf{0.36}   \\ \bottomrule
                            
\label{tab:2}
\end{tabularx}

\textit{Note: This table shows accuracy comparisons on the CONCEPTRULES V1 and V2 datasets, where a score of 1 indicates perfect accuracy. The comparison includes both simplified and full versions of the datasets. Notably, GPT-4 with `Zero-shot CoT' closely matches or occasionally surpasses our `ChatLogic' framework in performance. The best performance results for each version of the datasets are highlighted in bold.}
\end{table}

\section{Result}

The experimental findings, as illustrated in Tables \ref{tab:1} and \ref{tab:2}, unequivocally showcase the ChatLogic framework's significant enhancement of LLMs' performance, surpassing the Baseline with considerably higher accuracy in most scenarios. While native models demonstrate competency in answering straightforward questions, they exhibit limitations in complex multi-step reasoning tasks, reducing accuracy in more challenging questions. In stark contrast, the amalgamation of ChatLogic with LLMs consistently manifests superior accuracy across various levels of question difficulty. This highlights the critical role of augmenting LLMs with logical symbolic operations in multi-step reasoning. By adopting this methodology, we ensure the retention of comprehensive information in natural language, effectively preventing omissions and the accumulation of errors that could compromise reasoning outcomes. Moreover, this approach enhances the transparency of the reasoning process, thereby elevating the credibility and traceability of the results.

In our analysis of the PARARULE-Plus dataset, the ChatLogiour work's significant value and relevant model (`Base') and `Zero-shot CoT' in most scenarios. Notably, GPT-4, in conjunction with ChatLogic, exhibits exceptional performance on questions of higher complexity (Depth=4 and Depth=5), underscoring ChatLogic’s robust capability in handling intricate problems. Regarding Llama 2-7B, despite its weaker baseline performance, it shows significant improvement at all depth levels when assisted by ChatLogic. This indicates the framework's versatility in enhancing multi-step reasoning across different models.

Observing the CONCEPTRULES V1 and V2 datasets makes a notable shift in GPT-4’s performance evident. With Zero-shot CoT, GPT-4 either parallels or slightly surpasses ChatLogic in many cases, particularly in the full version of the CONCEPTRULES V2 dataset. The performance difference between GPT-4 and ChatLogic on these datasets is more nuanced than their performance on the PARARULE-Plus dataset. The inherent robustness of the GPT-4 model, likely due to its larger parameter count, already demonstrates formidable capabilities. This finding underscores the future need for more sophisticated datasets to challenge the upper-performance limits of advanced LLMs. Additionally, it's observed that ChatLogic primarily enhances models with a smaller parameter count by providing appropriate guidance, thereby boosting their performance on the task. This reaffirms our work's significant value and relevance, especially in optimizing models not inherently equipped with extensive computational resources.

\begin{table}[htbp]
\centering  
\caption{Test results of code executability across three datasets.}
\begin{tabular*}{\linewidth}{@{\extracolsep{\fill}}ccccc}
\toprule
Dataset                                              & Model      & Base  & SE   & SE+SYN \\ \midrule
\multicolumn{1}{c}{\multirow{3}{*}{\scriptsize CONCEPTRULES V1}} & \small GPT-3.5    & 0.63  & 0.68 & 0.7   \\
                                                     & \small GPT-4      & 0.92  & 0.96 & 0.96   \\  
                                                     & \small Llama 2-7B & 0.31  & 0.60 & 0.62  \\ \midrule
\multirow{3}{*}{\scriptsize CONCEPTRULES V2}                     & \small GPT-3.5    &  0.6     &  0.73    &  0.8      \\
                                                     & \small GPT-4      &  0.92     &  0.93    &  0.95      \\
                                                     & \small Llama 2-7B &  0.33     &  0.52    &  0.53      \\ \midrule
\multirow{3}{*}{\scriptsize PARARULE-Plus}                       & \small GPT-3.5    & 0.26     &  0.5    & 0.62      \\
                                                     & \small GPT-4      & 0.54      & 0.64     &  0.7       \\
                                                     & \small Llama 2-7B & 0.1       & 0.16     &  0.16       
\\\bottomrule
\label{tab:3}

\end{tabular*}

\textit{Note: Two modules respectively improve the executability of code, Semantic Correction (SE) and Semantic Correction + Syntax Correction (SE+SYN).}
\end{table}

\section{Ablation Study}

We have introduced two modules to improve code execution from semantic and syntax perspectives. To demonstrate their role in aiding LLMs' multi-step reasoning, we will separately assess how each module affects the code's executability. We anticipate a gradual increase in successful executions, which would validate the effectiveness of our approach.

For the PARARULE-Plus dataset, we increased our data sampling to 100 random samples from the entire dataset, ensuring that the selection was shuffled entirely, with no specific order applied to parameters such as `Depth (2-5)' and `Pattern (Animal \& People)'. For the CONCEPTRULES V1 and V2 datasets, we thoroughly shuffled all data from the `simplified' and `full' versions of each dataset and randomly selected 100 samples. To assess the executability rate of the generated code, we deployed it locally, evaluating solely based on the absence of error messages and the correctness of the output content.

The results are presented in Table \ref{tab:3}. In our comparison, we examined the enhancements in code execution rate achieved by LLMs with the gradual integration of different modules. Relative to the baseline, both modules demonstrated incremental improvements in execution rates. The Syntax Correction module still proved valuable despite Llama 2 not utilizing as much code text for pre-training as GPT-3.5 and GPT-4. While it didn't significantly increase the execution rate, its contribution to refining code quality is noteworthy. Furthermore, it's important to highlight that GPT-4, due to its extremely advanced capabilities in some subtasks, has seemingly reached a `performance ceiling' on the current datasets. This suggests that the dataset's limitations may somewhat constrain its potential.

\section{Limitation and Future Work}

Through experimental evaluations on multiple mainstream LLMs, we observed that ChatLogic+LLMs outperformed native LLMs in performance. The impressive performance demonstrates the effectiveness of our work. However, some issues have also been exposed. PARARULE-Plus is based on the closed-world assumption in question-answering data. Additionally, datasets like CONCEPTRULES V1 and V2, which are also artificially constructed, lack natural linguistic expression, which may not fully represent real-world complexities. When confronted with more complex sentences in the context of an open-world assumption, importing, integrating, and inferring external knowledge information that is expressed differently still poses challenges. Despite the valuable results from our experiments in enhancing code reliability, it's essential to acknowledge that the optimization module's applicability is currently limited to specific datasets. The carefully designed prompt sample cases are also optimized for specific data sets and are not a universal prompt template. Future developments should focus on creating adaptable optimization components \cite{marvie2005picolo} to address a wider array of scenarios and data sources.

\bibliography{conference_101719}

\end{document}